\documentclass[sigconf]{acmart}

\AtBeginDocument{%
  }

\copyrightyear{2026}
\acmYear{2026}
\setcopyright{cc}
\setcctype{by}
\acmConference[CIKM '26]{Proceedings of the 35th ACM International Conference on Information and Knowledge Management}{November 07--11, 2026}{Rome, Italy}
\acmBooktitle{Proceedings of the 35th ACM International Conference on Information and Knowledge Management (CIKM '26), November 07--11, 2026, Rome, Italy}
\acmDOI{10.1145/3799682.3839867}
\acmISBN{979-8-4007-2539-5/2026/11}

\usepackage[utf8]{inputenc}
\usepackage{url}
\usepackage{enumitem}
\usepackage{colortbl}
\usepackage{multirow}
\usepackage{bbding}
\newcommand{\hide}[1]{}
\begin{document}

\title{Can Large Language Models Identify Meaningful Touchpoints in Conversion Attribution?}

\author{Jinqi Wu}
\email{jinqiwu001@gmail.com}
\authornote{Equal contribution.}
\affiliation{%
  \institution{State Key Laboratory for Novel Software Technology, Nanjing University}
  \city{Nanjing}
  \country{China}}
\affiliation{%
  \institution{School of Intelligence Science and Technology, Nanjing University}
  \city{Suzhou}
  \country{China}}

\author{Sishuo Chen}
\authornotemark[1]
\email{chensishuo@pku.edu.cn}
\author{Zhangming Chan}
\email{zhangming.czm@alibaba-inc.com}
\author{Yong Bai}
\email{baiyong.by@alibaba-inc.com}
\affiliation{%
  \institution{Taobao \& Tmall Group of Alibaba}
  \city{Beijing}
  \country{China}}

\author{Chao Yi}
\email{yunan.yc@alibaba-inc.com}
\author{Han Zhu}
\email{zhuhan.zh@alibaba-inc.com}
\author{Shuodian Yu}
\email{yushuodian.ysd@alibaba-inc.com}
\affiliation{%
  \institution{Taobao \& Tmall Group of Alibaba}
  \city{Beijing}
  \country{China}}

\author{Lei Zhang}
\email{zl165646@alibaba-inc.com}
\author{Sheng Chen}
\email{chensheng.cs@alibaba-inc.com}
\author{Chenghuan Hou}
\email{jinyao@alibaba-inc.com}
\affiliation{%
  \institution{Taobao \& Tmall Group of Alibaba}
  \city{Beijing}
  \country{China}}

\author{Jian Xu}
\authornote{Corresponding author.}
\email{xiyu.xj@alibaba-inc.com}
\affiliation{%
  \institution{Taobao \& Tmall Group of Alibaba}
  \city{Beijing}
  \country{China}}

\author{Chaoyou Fu}
\email{bradyfu24@gmail.com}
\affiliation{%
  \institution{State Key Laboratory for Novel Software Technology, Nanjing University}
  \city{Nanjing}
  \country{China}}
\affiliation{%
  \institution{School of Intelligence Science and Technology, Nanjing University}
  \city{Suzhou}
  \country{China}}

\renewcommand{\shortauthors}{Jinqi Wu et al.}

\newcommand{\acronym}[1]{\underline{\textbf{#1}}}

\begin{abstract}
Touchpoint selection in conversion attribution, namely identifying meaningful touchpoints contributing to conversions, is essential for e-commerce recommendation and online advertising.
Current selection methods rely heavily on collaborative-filtering-based heuristics, which fail to align with user-perceived semantic intent. 
Through human annotation, we reveal a significant \textbf{semantic gap}: many \textbf{implicitly-related, semantically relevant touchpoints} remain undetected by existing rules.
Therefore, we systematically evaluate the capability of Large Language Models (\textbf{LLMs}) in identifying these hidden associations. 
Our evaluation shows that while LLMs effectively uncover a substantial portion of implicitly-related touchpoints, significant room for improvement remains in their selection performance. 
Furthermore, we analyze the impact of different prompting strategies and foundation model choices on identification performance, providing valuable insights into their reasoning patterns and effectiveness.
These insights offer a new roadmap for transitioning conversion attribution from mechanical rule-matching to human-aligned semantic reasoning.
Moreover, we leverage the LLM-attributed conversion labels for enhancing industrial CVR model training and achieve significant offline performance gains, showing the potential of LLMs in conversion attribution.
\end{abstract}



\maketitle


\section{Introduction}
\begin{figure}[t]
    \centering
    \includegraphics[width=0.98\columnwidth]{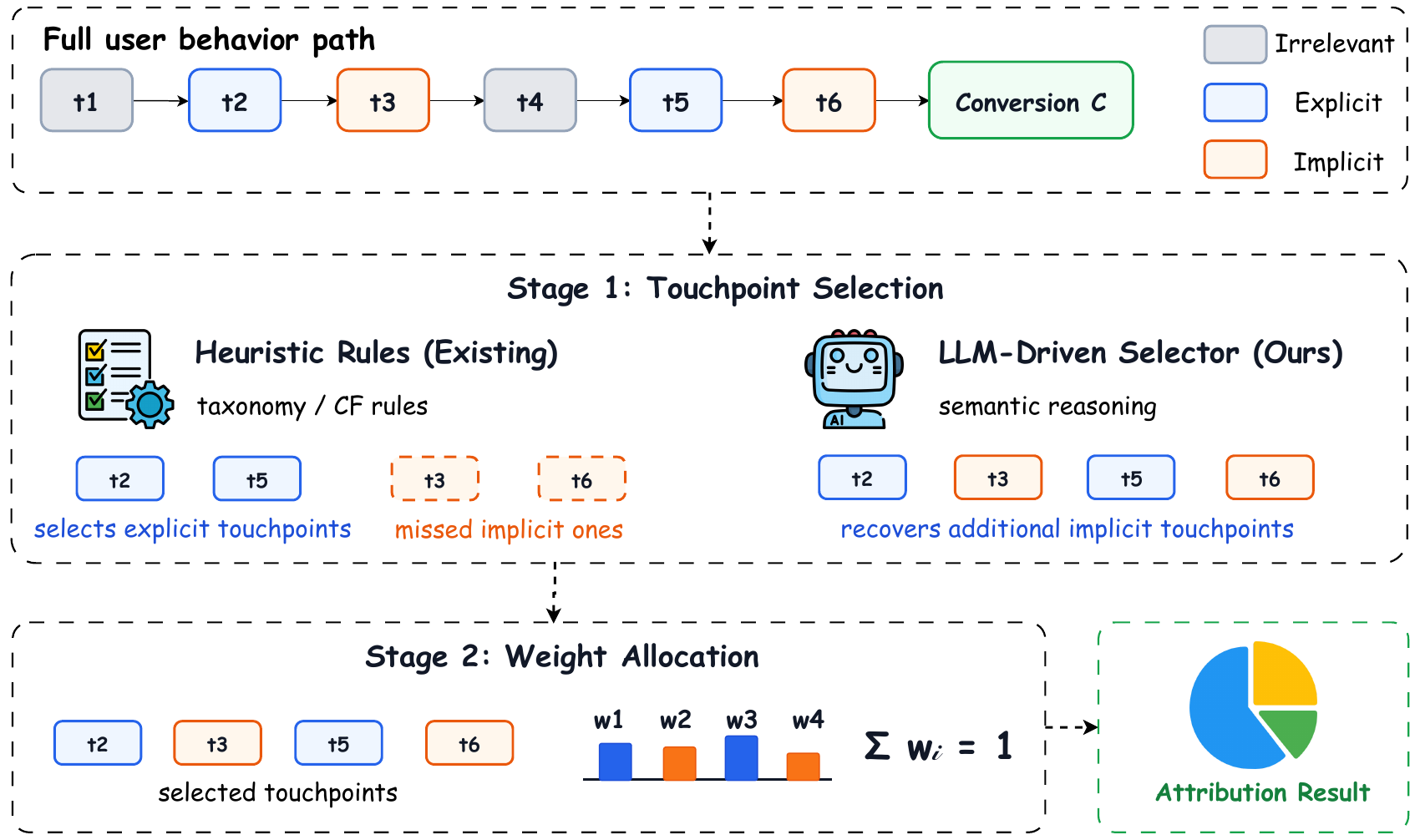}
    \caption{Touchpoint selection and weight allocation in conversion attribution.}
    \Description{Overview of conversion attribution with touchpoint selection followed by weight allocation.}
    \label{fig:attribution-overview}
    \vspace{-0.2cm}
\end{figure}

Conversion attribution, namely the mechanism allocating credits for a conversion across past user touchpoints, is a cornerstone for e-commerce platforms, which affects ad performance presentation to advertisers~\cite{shao2011data,kumar2020camta, yao2022causalmta,liu2026almmta}, label generation for conversion rate (CVR) prediction models~\cite{chen2025see,zeng2025clickabuyb,wu2026mac}, and ultimately, platform revenue.
Past studies on conversion attribution primarily focus on \textbf{(1) weight allocation} given a sequence of user touchpoints, which is usually restricted to the touchpoints under \textbf{the same item or shop} as the consumed product~\cite{yao2022causalmta,chen2025see}.
Recently, \citet{zeng2025clickabuyb} from Meta found that \textbf{(2) touchpoint selection} is another important aspect of conversion attribution, which aims to identify meaningful touchpoints from \textbf{the full user  behavior sequence}.
Figure~\ref{fig:attribution-overview} illustrates the two key components of conversion attribution.

Although \citet{zeng2025clickabuyb} revealed the value of full-path attribution for mining conversion signals and enhancing CVR prediction model training, their touchpoint selection relies on \textbf{heuristic rules} based on product taxonomy and collaborative-filtering (CF) signals,  which may fail to capture \textbf{implicitly-related touchpoints} that are semantically related to the target product but lack explicit taxonomy or historical engagement overlaps.
To verify this, we construct \textbf{SILVA}, the first benchmark for touchpoint \acronym{S}election \acronym{I}n fu\acronym{L}l-path con\acronym{V}ersion \acronym{A}ttribution, which is collected from a large-scale e-commerce platform.
Human annotation results on \textbf{SILVA} reveal a striking fact: beyond the \textbf{explicitly-related} touchpoints covered by taxonomy and CF-based heuristics, \emph{there exists a large number of \textbf{implicitly-related} touchpoints that are currently overlooked}.
These findings underscore a critical deficiency in current conversion attribution approaches, which remain blind to a vast space of implicitly-related, semantically relevant touchpoints.

To bridge this gap and explore how to identify these implicitly-related touchpoints, we turn to Large Language Models (LLMs) for their superior semantic reasoning capabilities~\cite{wei2022chain}. 
Specifically, we conduct a systematic evaluation of representative LLMs on SILVA, assessing their effectiveness in identifying both explicitly and implicitly-related touchpoints across various model scales and prompting strategies.
We summarize the main findings as follows:
\begin{enumerate}[leftmargin=*]
    \item \textbf{LLM Capability of Touchpoint Selection}: Cutting-edge LLMs are excellent at identifying explicitly-related touchpoints and effectively discover a substantial number of implicitly-related touchpoints, but  substantial room for improvement remains in implicitly-related touchpoint selection.
    \item \textbf{Effect of Prompting Strategies}: Pairwise reasoning outperforms listwise reasoning, but the gap narrows when model capability increases.
    \item \textbf{Performance across LLM Families}: Proprietary LLMs, such as Gemini 3.1 Pro~\cite{googledeepmind2026gemini31pro} and GPT 5.5~\cite{openai2026gpt55systemcard}, only slightly outperform open-weight counterparts, such as GLM 5.1~\cite{zeng2026glm5} and DeepSeek V4-Pro~\cite{deepseekai2026deepseekv4}; smaller LLMs (<100B) largely lag behind flagship models.
\end{enumerate}

Furthermore, inspired by these insights, we leverage conversion labels attributed by LLMs as auxiliary signals for training CVR prediction models, following \citet{chen2025see} and \citet{zeng2025clickabuyb}.
Results show that our approach, \acronym{L}LM-\acronym{O}riented \acronym{T}ouchpoint \acronym{U}nderstanding and \acronym{S}election (\textbf{LOTUS}), yields an absolute GAUC gain of \textbf{0.35 percentage points (pp)} over the production baseline (Base)~\cite{chen2025see} and a further gain of \textbf{0.15 pp} over the heuristic CABB competitor~\cite{zeng2025clickabuyb} for predicting the main conversion goal on a large-scale e-commerce platform with hundreds of millions of active users.
We believe that our systematic analysis and findings open up a new avenue for harnessing LLMs for touchpoint selection in conversion attribution, which contributes to growth in user satisfaction and platform revenue for e-commerce platforms.

\section{Related Work}

Conversion attribution mechanisms, namely the rules allocating conversion credits among user touchpoints, play a vital role in e-commerce platforms~\cite{shao2011data}.
Prior studies on conversion attribution focus on weight allocation among touchpoints via causal modeling~\cite{shao2011data,zhou2019deep,du2019causally,kumar2020camta, yao2022causalmta,bencina2025lidda,lewis2025amazon, liu2026almmta} and leveraging attribution results for training CVR prediction models~\cite{chen2025see,zeng2025clickabuyb,wu2026mac}.
However, almost all of them ignore \textbf{touchpoint selection} and only consider the touchpoints under the same item or shop as the consumed product  by default.
\citet{zeng2025clickabuyb} proposed to mine cross-item relevant touchpoints in a heuristic way based on collaborative-filtering signals and showed the value of touchpoint selection for enhancing the CVR prediction model training at Meta.
Nonetheless, their simple heuristic selection strategies may overlook implicitly-related touchpoints.
To our knowledge, our study takes \emph{the first step to investigate the characteristics of implicitly-related touchpoints} and explore how to \emph{harness LLMs for touchpoint selection} in conversion attribution.

\section{The SILVA Benchmark}

\subsection{Problem Formulation}

\textbf{Touchpoint selection} aims to identify whether a historical interaction contributed to a conversion. 
Formally, given a conversion $C$ and its preceding touchpoints $\{t_1, t_2, \dots, t_n\}$, we assign a label $l_i \in \{ \text{Irrelevant, Explicit, Implicit} \}$ to each $t_i$.
In the e-commerce context, meaningful touchpoints are categorized as:
\begin{itemize}
    \item \textbf{Explicitly-Related}: Interactions sharing metadata-driven or statistical associations with $C$: (1) \textit{Taxonomy Matching}, such as identical item, shop, brand, or leaf category; (2) \textit{CF-based Similarity}, involving similar leaf categories determined by collaborative-filtering signals~\cite{zeng2025clickabuyb}. 
    \item \textbf{Implicitly-Related}: Touchpoints with semantic-driven associations that elude explicit rules, including: (1) \textit{Functional Complementarity} (e.g., camera and memory card); and (2) \textit{Scenario/Audience Alignment} (e.g., shared anime IPs).
\end{itemize}


\subsection{Benchmark Construction and Analysis}

We establish \textbf{SILVA}, the first benchmark for \acronym{S}election \acronym{I}n fu\acronym{L}l-path con\acronym{V}ersion \acronym{A}ttribution. 
Collected from an e-commerce platform with hundreds of millions of daily active users, \textbf{SILVA} comprises $1,000$ unique conversion events and their preceding touchpoints within a 3-day lookback window, totaling $69,680$ touchpoints. 
Each touchpoint is enriched with multimodal information about the clicked product, including its image, title, shop, and category metadata.

\begin{figure}[t]
    \centering
    \includegraphics[width=0.98\columnwidth]{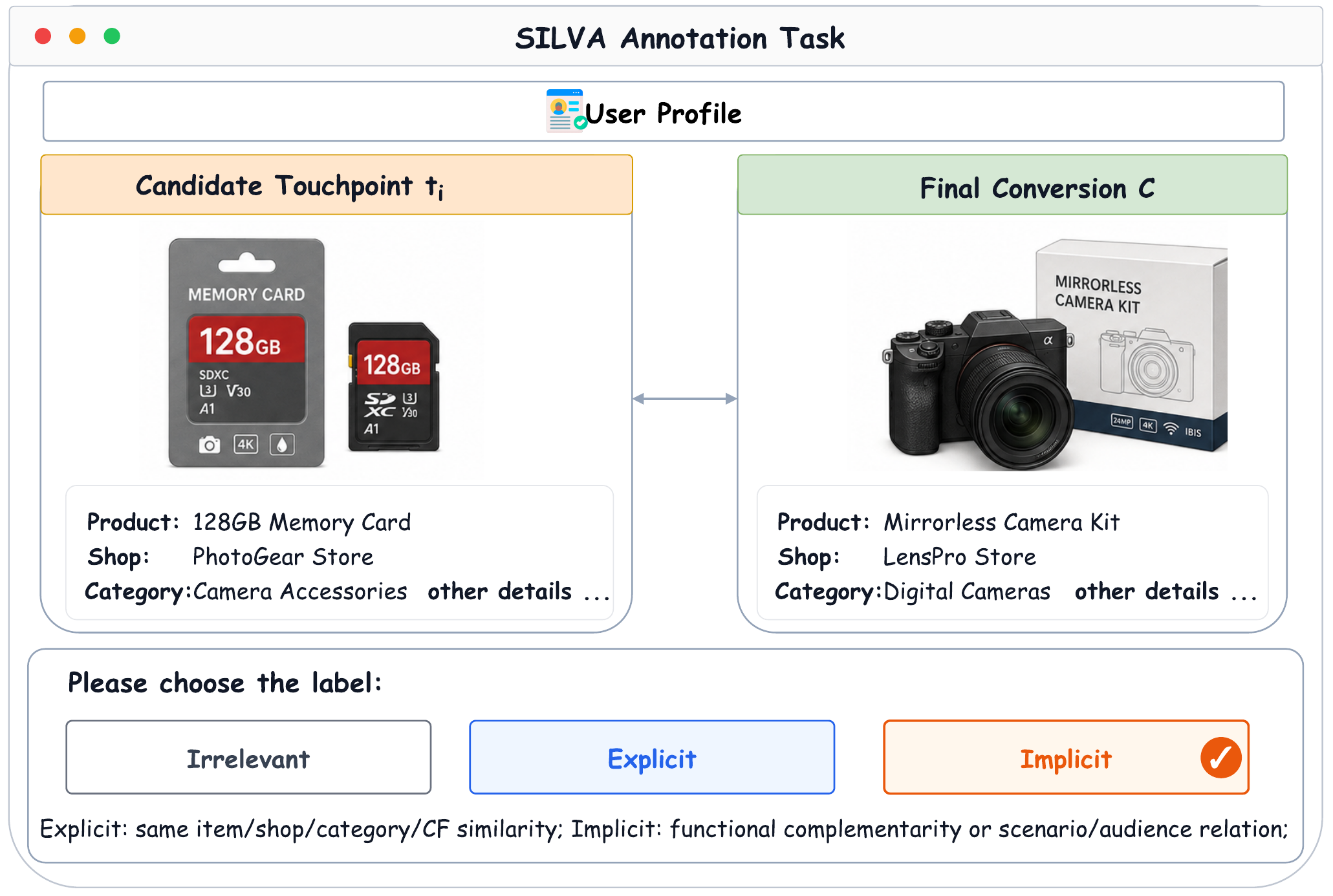}
    \caption{Schematic annotation interface for constructing the SILVA benchmark.} 
    \Description{A schematic annotation interface showing a user profile, a candidate touchpoint, the final conversion, and label choices for irrelevant, explicit, and implicit relevance.}
    \label{fig:silva-annotation}
    \vspace{-0.5cm}
\end{figure}

\textbf{Annotation reliability.} For human annotation, we recruited six well-trained annotators to assign labels using the annotation interface shown in Figure~\ref{fig:silva-annotation}.
All annotators were fairly compensated according to local prevailing wage rates.
Before formal annotation, annotators were trained with detailed guidelines and representative examples for explicit and implicit relevance. Importantly, annotators assigned labels solely according to the semantic annotation guidelines and were not shown the outputs of the taxonomy- or CF-based heuristic baselines. Each touchpoint was independently labeled by three annotators sampled from the pool of six annotators. We measure inter-annotator agreement using Fleiss' $\kappa$ over the three-way label space, i.e., Irrelevant, Explicit, and Implicit. The annotations achieve Fleiss' $\kappa>0.70$, indicating substantial agreement beyond chance; the final label is determined by majority voting among the three annotations.

Results reveal that explicitly-related touchpoints account for 19.09\% of the dataset.
Crucially, \textbf{implicitly-related touchpoints constitute a substantial 14.91\%}, which elude traditional taxonomy and CF-based rules. 
These statistics underscore that \textbf{implicit signals are non-negligible} and represent a significant portion of the user’s conversion journey currently overlooked by existing studies on conversion attribution~\cite{yao2022causalmta,chen2025see,zeng2025clickabuyb}.
To fill in the gap, we investigate the capability of LLMs in discovering these latent associations in \S~\ref{sec:exp} and demonstrate the downstream value of these implicitly-related touchpoints for enhancing CVR prediction in \S~\ref{sec:app}.



\section{Harnessing LLMs for Touchpoint Selection}
\label{sec:exp}

\begin{table*}[t]
    \centering
    \caption{Main results of LLM-based touchpoint selection on SILVA. Exp. and Imp. denote explicit and implicit touchpoints. Shaded columns mark F1 metrics, with bold and underline indicating the best and second-best values in each F1 column.}
    \label{tab:llm-main-results}
    \normalsize
    \setlength{\tabcolsep}{1.4pt}
    \begin{tabular*}{0.925\textwidth}{@{\extracolsep{\fill}}l%
        cc>{\columncolor{black!8}}c%
        cc>{\columncolor{black!8}}c%
        >{\columncolor{black!8}}c%
        cc>{\columncolor{black!8}}c%
        cc>{\columncolor{black!8}}c%
        >{\columncolor{black!8}[\tabcolsep][0pt]}c@{}}
        \toprule
        \multirow{3.5}{*}{\textbf{Model}} & \multicolumn{7}{c}{\textbf{Listwise Prompting}} & \multicolumn{7}{c}{\textbf{Pairwise Prompting}} \\
        \cmidrule(lr){2-8} \cmidrule(lr){9-15}
        & \multicolumn{3}{c}{\textbf{Exp.}} & \multicolumn{3}{c}{\textbf{Imp.}} & \multicolumn{1}{c}{\textbf{Overall}}
        & \multicolumn{3}{c}{\textbf{Exp.}} & \multicolumn{3}{c}{\textbf{Imp.}} & \multicolumn{1}{c}{\textbf{Overall}} \\
        \cmidrule(lr){2-4} \cmidrule(lr){5-7} \cmidrule(lr){9-11} \cmidrule(lr){12-14}
        & \textbf{P} & \textbf{R} & \multicolumn{1}{c}{\textbf{F1}} & \textbf{P} & \textbf{R} & \multicolumn{1}{c}{\textbf{F1}} & \multicolumn{1}{c}{\textbf{F1}} & \textbf{P} & \textbf{R} & \multicolumn{1}{c}{\textbf{F1}} & \textbf{P}& \textbf{R}& \multicolumn{1}{c}{\textbf{F1}} & \multicolumn{1}{c}{\textbf{F1}} \\
        \midrule
        \multicolumn{15}{l}{\textit{Closed-source large models}} \\
        GPT-5.5 & 0.9180 & 0.9418 & \textbf{0.9070} & 0.5161 & 0.4763 & 0.4732 & \underline{0.6901} & 0.8747 & 0.9751 & 0.8968 & 0.6454 & 0.5567 & \textbf{0.5426} & \textbf{0.7197} \\
        Gemini-3.1-Pro & 0.9312 & 0.9275 & \underline{0.9053} & 0.5612 & 0.5201 & \textbf{0.5077} & \textbf{0.7065} & 0.8953 & 0.9686 & \textbf{0.9095} & 0.6246 & 0.5398 & \underline{0.5293} & \underline{0.7194} \\
        Claude-Opus-4.7 & 0.9151 & 0.9374 & 0.9002 & 0.4698 & 0.4664 & 0.4491 & 0.6747 & 0.9261 & 0.9147 & 0.8975 & 0.5707 & 0.6210 & 0.4979 & 0.6977 \\
        \midrule
        \multicolumn{15}{l}{\textit{Open-weight large models}} \\
        GLM-5.1 & 0.9074 & 0.8186 & 0.8148 & 0.5748 & 0.5050 & 0.4668 & 0.6408 & 0.9116 & 0.9458 & \underline{0.9092} & 0.6108 & 0.5576 & 0.5080 & 0.7086 \\
        DeepSeek-V4-Pro & 0.9044 & 0.8480 & 0.8448 & 0.5472 & 0.5577 & \underline{0.4990} & 0.6719 & 0.9115 & 0.9250 & 0.8956 & 0.6292 & 0.5474 & 0.5155 & 0.7056 \\
        DeepSeek-V4-Flash & 0.9117 & 0.8396 & 0.8415 & 0.5185 & 0.4948 & 0.4694 & 0.6554 & 0.9249 & 0.9109 & 0.8975 & 0.5693 & 0.4761 & 0.4586 & 0.6780 \\
        \midrule
        \multicolumn{15}{l}{\textit{Open-weight smaller models}} \\
        Qwen3.5-27B & 0.9292 & 0.7636 & 0.7804 & 0.4772 & 0.4436 & 0.4232 & 0.6018 & 0.9345 & 0.8875 & 0.8852 & 0.4969 & 0.4541 & 0.4420 & 0.6636 \\
        Qwen3.5-30B-A3B & 0.9317 & 0.6608 & 0.7034 & 0.4753 & 0.4437 & 0.3983 & 0.5509 & 0.9013 & 0.9327 & 0.8915 & 0.5325 & 0.4713 & 0.4491 & 0.6703 \\
        \bottomrule
    \end{tabular*}
\end{table*}


\subsection{Experimental Setup}

\textbf{Foundation models.}
We evaluate eight representative LLMs: GPT-5.5~\cite{openai2026gpt55systemcard}, Gemini-3.1-Pro~\cite{googledeepmind2026gemini31pro}, Claude-Opus-4.7~\cite{anthropic2026claudeopus47}, GLM-5.1~\cite{zeng2026glm5}, DeepSeek-V4-Pro and DeepSeek-V4-Flash~\cite{deepseekai2026deepseekv4}, and Qwen3.5-27B and Qwen3.5-30B-A3B~\cite{qwen2026qwen35}. For models supporting switching thinking on or off, we use the thinking mode; otherwise, we use the default setting. 

\begin{figure}[t]
    \centering
    \includegraphics[width=0.95\columnwidth]{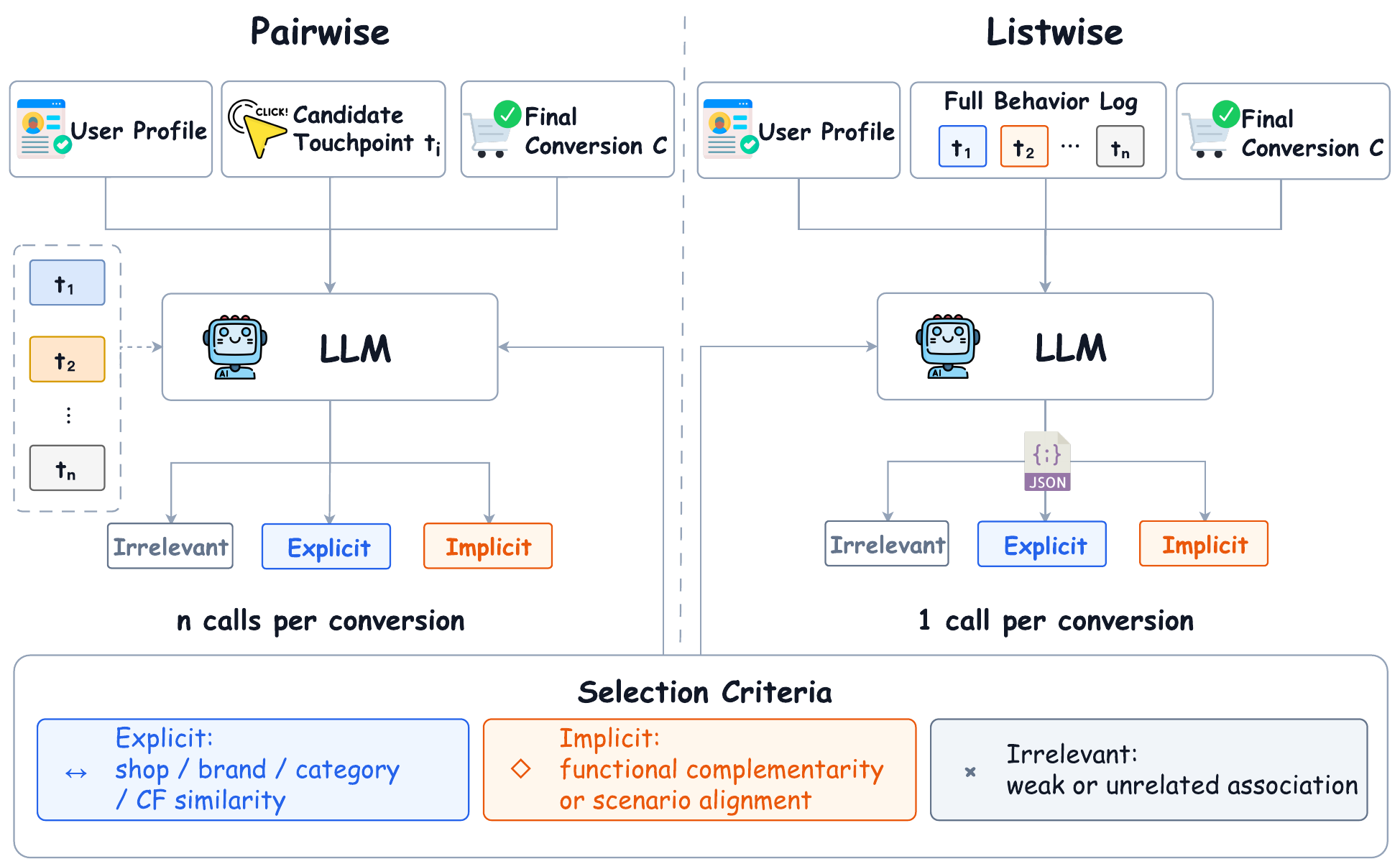}
    \caption{Schematic comparison of pairwise and listwise prompting protocols for LLM-based touchpoint selection.}
    \Description{A schematic comparison of pairwise and listwise prompting protocols, showing the inputs, LLM calls, output labels, and shared selection criteria.}
    \label{fig:prompting-protocols}
    \vspace{-0.2cm}
\end{figure}

\textbf{Prompting protocols.}
We compare pairwise and listwise prompting. 
In the \textbf{pairwise} protocol, the model receives the user profile, one candidate touchpoint, and the final conversion, and judges whether the touchpoint is irrelevant, explicitly related, or implicitly related. In the \textbf{listwise} protocol, the model receives the user profile, the full behavior log, and the final conversion, and returns the relevant touchpoints with their relation types. 
Figure~\ref{fig:prompting-protocols} compares the two protocols.
The same criteria are listed in the prompts in both protocols: explicit relevance is based on shop/brand matching, leaf-category matching, and CF-based similar-category matching, while implicit relevance covers functional complementarity and scenario or audience consistency.

\textbf{Evaluation metrics.}
We evaluate each conversion separately and then macro-average across conversions. For each conversion $j$ and class $c \in \{\mathrm{Exp.}, \mathrm{Imp.}\}$, we compute the precision $P_{c,j}$, the recall $R_{c,j}$, and the F1 score $F1_{c,j}=2P_{c,j}R_{c,j}/(P_{c,j}+R_{c,j})$ from the touchpoints in that conversion.
Table~\ref{tab:llm-main-results} reports the averaged $\overline{P}_c$, $\overline{R}_c$, and $\overline{F1}_c$. The reported Overall F1 is computed as $(\overline{F1}_{\mathrm{Exp.}}+\overline{F1}_{\mathrm{Imp.}})/2$.

\vspace{-0.2cm}

\subsection{Main Results and Key Findings}

We present the main results in Table~\ref{tab:llm-main-results} and make four key findings.

\textbf{Finding 1: Implicitly-related touchpoints are much harder to identify than explicitly-related ones.}
Across all configurations, explicit F1 is consistently higher than implicit F1. Pairwise prompting reaches a high average explicit F1 at 0.8979, but only achieves 0.4929 implicit F1; listwise prompting averages 0.8372 explicit F1 and 0.4608 implicit F1. The best implicit F1 is 0.5426, far below the best explicit F1 of 0.9095, indicating that semantic relations such as functional complementarity and scenario consistency remain difficult even for strong LLMs.
Considering that the expected F1 score for random guessing is approximately $0.1491$ (based on the prevalence of implicit labels in SILVA), \emph{the performance of current LLMs is indeed significant, yet a substantial gap remains toward human-level understanding}.

\textbf{Finding 2: Pairwise prompting beats listwise prompting.}
Pairwise prompting outperforms listwise prompting on every model in Table~\ref{tab:llm-main-results}.
Averaged over all models, pairwise prompting raises the overall F1 from 0.6490 to 0.6954.
Pairwise prompting likely benefits from decomposing a noisy sequence-level selection problem into candidate-level judgments, while listwise prompting tends to over-filter touchpoints, resulting in a significantly lower recall. 


\textbf{Finding 3: Strong open-weight models are competitive with proprietary counterparts, but smaller open-weight variants lag behind.}
GLM-5.1 and DeepSeek-V4-Pro reach 0.7086 and 0.7056 in terms of overall F1 under pairwise prompting, respectively, close to the best proprietary model GPT-5.5 at 0.7197. 
In contrast, smaller Qwen variants lag behind flagship models, especially on the implicitly-related touchpoints. 
This performance gap underscores that capturing latent semantic associations for touchpoint selection necessitates the superior reasoning power of large flagship models.

\begin{figure}[t]
     \centering
     \includegraphics[width=0.9\columnwidth]{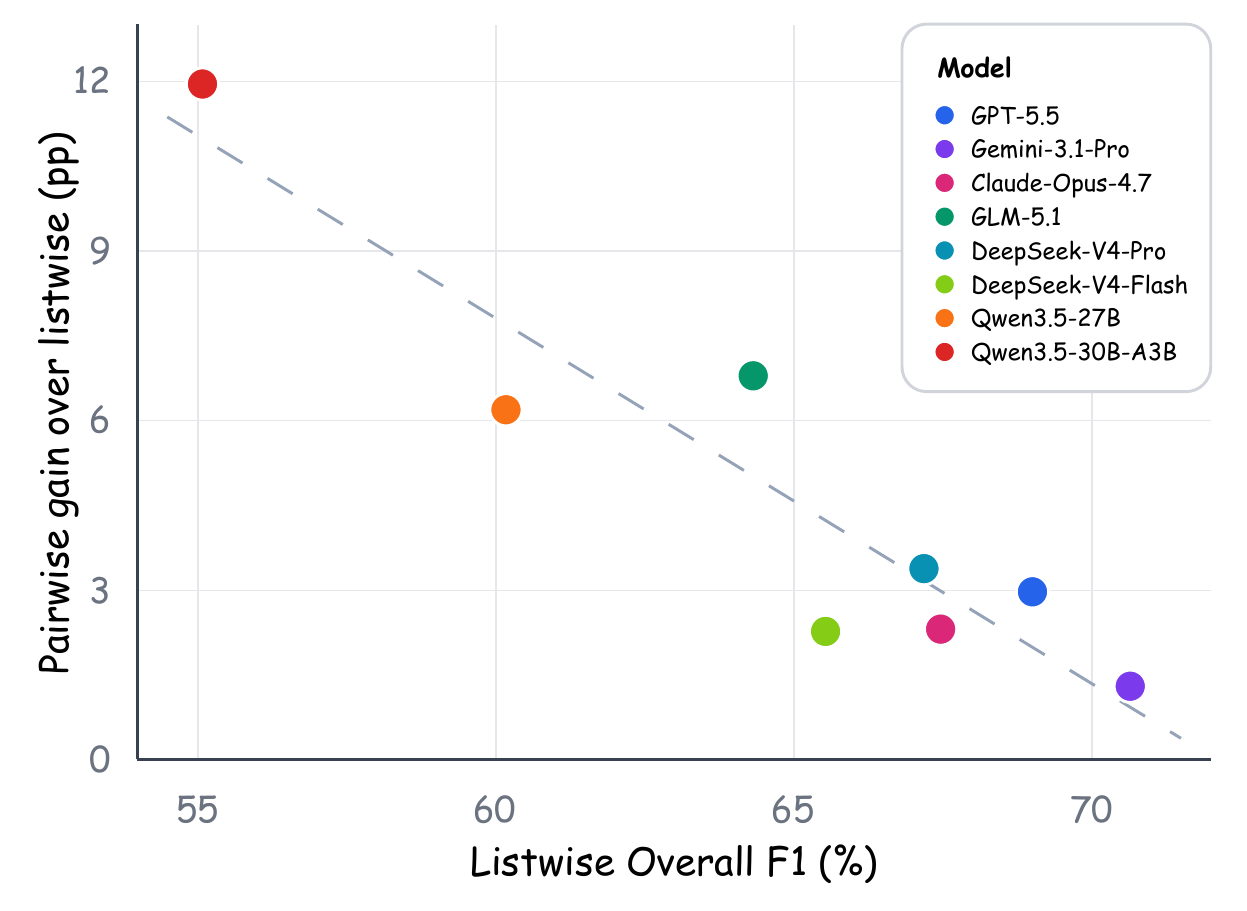}
     \caption{The performance gap between pairwise and listwise prompting narrows when the listwise overall F1 is higher.}
     \Description{A scatter plot showing listwise Overall F1 on the horizontal axis and pairwise gain over listwise prompting on the vertical axis for the evaluated LLMs.}
     \label{fig:protocol-gap}
     \vspace{-0.2cm}
 \end{figure}

\textbf{Finding 4: Model capability narrows the performance gap between pairwise and listwise prompting strategies.}
The gap is 1.3 percentage points (\textbf{pp} for short) for Gemini-3.1-Pro and 3.0 pp for GPT-5.5, but grows to 6.2 pp for Qwen3.5-27B and 11.9 pp for Qwen3.5-30B-A3B.
This suggests that stronger models handle listwise comparison better, whereas smaller models benefit more from pairwise decomposition. 
Figure~\ref{fig:protocol-gap} further shows that the pairwise gain is larger when listwise Overall F1 is lower.
This trend indicates that pairwise prompting is especially useful when a model struggles to compare many touchpoints in a single listwise response.

\subsection{Ablation Study}

We further analyze the impact of chain-of-thought reasoning by switching the reasoning mode from \texttt{think} to \texttt{nothink} and comparing the resulting metrics in Table~\ref{tab:reasoning-ablation}.
We find that the \textit{thinking mode is generally helpful, but its effect varies across models and protocols}.
On average, it brings a larger Overall F1 gain under listwise prompting (+0.0196) than under pairwise prompting (+0.0082), while the strongest listwise gains come from the smaller open-weight Qwen variants.
Overall, the thinking mode provides only marginal improvements, suggesting that current reasoning patterns are not fully optimized for the complexities of touchpoint selection. This indicates a critical need for future research to develop specialized fine-tuning or reasoning strategies specifically adapted to the nuanced semantic judgments required for the touchpoint selection task.~\looseness=-1
 
\begin{table}[t] \small
    \centering
    \caption{Ablation of thinking mode for LLM-based touchpoint selection. We report the F1 gain of \texttt{think} over \texttt{nothink}.}
    \label{tab:reasoning-ablation}
    \normalsize
    \begin{tabular*}{\columnwidth}{@{\extracolsep{\fill}}llccc@{}}
        \toprule
        \textbf{Prompt} &\textbf{ Model} & \textbf{$\Delta$Exp.} &\textbf{$\Delta$Imp.} & \textbf{$\Delta$Overall} \\
        \midrule
        \multirow{7.5}{*}{Listwise}& GPT-5.5 & +0.0165 & +0.0296 & +0.0231 \\
         & Gemini-3.1-Pro & +0.0089 & +0.0089 & +0.0089 \\
        & DeepSeek-V4-Pro & +0.0036 & +0.0016 & +0.0026 \\
        & DeepSeek-V4-Flash & -0.0012 & +0.0063 & +0.0025 \\
         & Qwen3.5-27B & +0.0826 & +0.0239 & +0.0533 \\
         & Qwen3.5-30B-A3B & +0.0452 & +0.0094 & +0.0274 \\
         & \textit{Avg.} & +0.0259 & +0.0133 & +0.0196 \\
        \midrule
  \multirow{7.5}{*}{Pairwise}        & GPT-5.5 & -0.0010 & +0.0014 & +0.0002 \\
         & Gemini-3.1-Pro & +0.0010 & +0.0157 & +0.0084 \\
         & DeepSeek-V4-Pro & +0.0033 & +0.0156 & +0.0095 \\
         & DeepSeek-V4-Flash & -0.0014 & +0.0140 & +0.0062 \\
         & Qwen3.5-27B & +0.0028 & +0.0022 & +0.0025 \\
         & Qwen3.5-30B-A3B & +0.0313 & +0.0130 & +0.0222 \\
         & \textit{Avg.} & +0.0060 & +0.0103 & +0.0082 \\
        \bottomrule
    \end{tabular*}
    \vspace{-0.2cm}
\end{table}

\section{Application: Enhancing CVR Prediction}
\label{sec:app}

\begin{table}[t]  \small
\centering
\caption{Comparison of different touchpoint selection methods for CVR prediction. Our LLM-based selection captures incremental semantic signals beyond traditional heuristics.}  \label{table:gauc}
\resizebox{0.475\textwidth}{!}{
\begin{tabular}{lcccc} 
\toprule \textbf{Method} & \textbf{In-Shop} & \textbf{Heuristic} & \textbf{LLM-Based} & \textbf{$\Delta$GAUC(pp)} \\ \midrule Base~\cite{chen2025see} & \Checkmark & \XSolidBrush & \XSolidBrush & - \\ CABB \cite{zeng2025clickabuyb} & \Checkmark & \Checkmark & \XSolidBrush & +0.20 \\ \textbf{LOTUS (Ours)} & \Checkmark & \Checkmark & \Checkmark & \textbf{+0.35} \\ \bottomrule \end{tabular}}
\end{table}

To validate the potential utility of LLM-based touchpoint selection, we conduct an experiment using a production-scale CVR prediction model on our e-commerce platform.
Following \citet{chen2025see} and \citet{zeng2025clickabuyb}, we incorporate the selected touchpoints as positive samples for \textbf{an auxiliary indirect conversion prediction task}.
Leveraging the insights from our benchmark evaluation in \S~\ref{sec:exp}, we employ our internal foundation model with pairwise reasoning to perform touchpoint selection at scale.

As shown in Table~\ref{table:gauc}, our approach, \acronym{L}LM-\acronym{O}riented \acronym{T}ouchpoint \acronym{U}nderstanding and \acronym{S}election (\textbf{LOTUS}), substantially outperforms both the in-shop attribution baseline Base~\cite{chen2025see} and the heuristic rule-based selection method CABB~\cite{zeng2025clickabuyb}, which covers only explicit cross-shop touchpoints, achieving absolute offline Group AUC (GAUC) gains of \textbf{0.35 pp and 0.15 pp}~\cite{hu2023ps,chen2025see,luo2026modeling,liu2026est,li2026delayed} for the main conversion goal over Base and CABB, respectively.
The improvement over Base demonstrates the value of incorporating cross-shop touchpoints, while the additional gain over CABB suggests that LLM-based semantic reasoning can identify useful implicit relations beyond those captured by explicit heuristic rules.

Considering that an absolute GAUC improvement of $0.1$ pp is empirically sufficient to drive substantial growth in online sales and advertising revenue~\cite{chen2025see,chan2023capturing,zhang2022keep}, these results validate the practical utility of leveraging LLMs to discover latent, high-value touchpoints that traditional methods overlook.

\section{Conclusion}

In this study, we identify a critical \textbf{semantic gap} in conversion attribution, where traditional heuristic methods overlook a vast space of \textbf{implicitly-related touchpoints}.
Through our newly established \textbf{SILVA} benchmark, we find that implicitly-related and explicitly-related touchpoints account for 14.91\% and 19.09\% of all annotated touchpoints, respectively.
Our systematic evaluation shows that LLMs can effectively bridge this gap by reasoning over product metadata and user-behavior context.
Detailed analysis further reveals that pairwise reasoning improves performance and that larger foundation models are better able to capture subtle semantic nuances.
Furthermore, integrating these LLM-identified touchpoints into a production-scale CVR prediction model improves GAUC by an absolute margin of \textbf{0.35 pp} over the production baseline~\cite{chen2025see}, which considers only in-shop touchpoints, and by \textbf{0.15 pp} over the heuristic CABB competitor~\cite{zeng2025clickabuyb}.
Overall, our study highlights the promising potential of LLMs in refining conversion attribution, providing a useful roadmap for future research on conversion attribution and conversion rate prediction in the LLM era.

\begin{acks}
This work is funded by CCF-ALIMAMA TECH Kangaroo Fund (No. CCF-ALIMAMA of 2025004).
\end{acks}

\section*{GenAI Usage Disclosure}
Generative AI tools were only used for limited text polishing during the draft preparation.

\bibliographystyle{ACM-Reference-Format}
\balance
\bibliography{references}

\end{document}